\documentclass[letterpaper, 10pt, conference]{ieeeconf}
\IEEEoverridecommandlockouts
\usepackage{cite}
\usepackage{amsmath,amssymb,amsfonts}
\usepackage{algorithmic}
\usepackage{graphicx}
\usepackage{textcomp}
\usepackage{xcolor}
\usepackage{subcaption}
\usepackage{pifont}

\usepackage[utf8]{inputenc} % allow utf-8 input
\usepackage[T1]{fontenc}    % use 8-bit T1 fonts
\usepackage{hyperref}       % hyperlinks
\usepackage{url}            % simple URL typesetting
\usepackage{booktabs}       % professional-quality tables
\usepackage{amsfonts}       % blackboard math symbols
\usepackage{nicefrac}       % compact symbols for 1/2, etc.
\usepackage{microtype}      % microtypography
\usepackage{xcolor}
\usepackage{tabularx}
\usepackage{dblfloatfix} 
\usepackage{array}
\usepackage{multirow}
\usepackage{float}
\usepackage{pifont}
\usepackage{adjustbox}
\usepackage[compatibility=false]{caption}
\usepackage{subcaption}
\usepackage{cuted}
\newcommand{\cmark}{\ding{51}}%
\newcommand{\xmark}{\ding{55}}%

\newcommand{\starfull}{\ding{72}}   
\newcommand{\starempty}{\ding{73}} 
\newcommand{\fivestars}[2][-.05em]{%
  \def\gap{\kern#1}%
  \ifnum#2>0 \starfull\gap\else \starempty\gap\fi
  \ifnum#2>1 \starfull\gap\else \starempty\gap\fi
  \ifnum#2>2 \starfull\gap\else \starempty\gap\fi
  \ifnum#2>3 \starfull\gap\else \starempty\gap\fi
  \ifnum#2>4 \starfull\else \starempty\fi
}

\begin{document}

\title{\LARGE \bf Humanoid Everyday: A Comprehensive Robotic Dataset \\for Open-World Humanoid Manipulation}

\author{
Zhenyu Zhao\textsuperscript{1*}
Hongyi Jing\textsuperscript{1*}\thanks{* Equal contribution. $^{\dag}$ Equal advising.}
Xiawei Liu\textsuperscript{1}
Jiageng Mao\textsuperscript{1$\dag$} \\
Abha Jha\textsuperscript{1}
Hanwen Yang\textsuperscript{1}
Rong Xue\textsuperscript{1}
Sergey Zakharov\textsuperscript{2}
Vitor Guizilini\textsuperscript{2}
Yue Wang\textsuperscript{1$\dag$} \\
\textsuperscript{1}University of Southern California \quad
\textsuperscript{2}Toyota Research Institute}
% \textsuperscript{*}Equal Contribution}

\maketitle

\begin{strip}
  \vspace{-12mm}
  \centering
  \includegraphics[width=\textwidth]{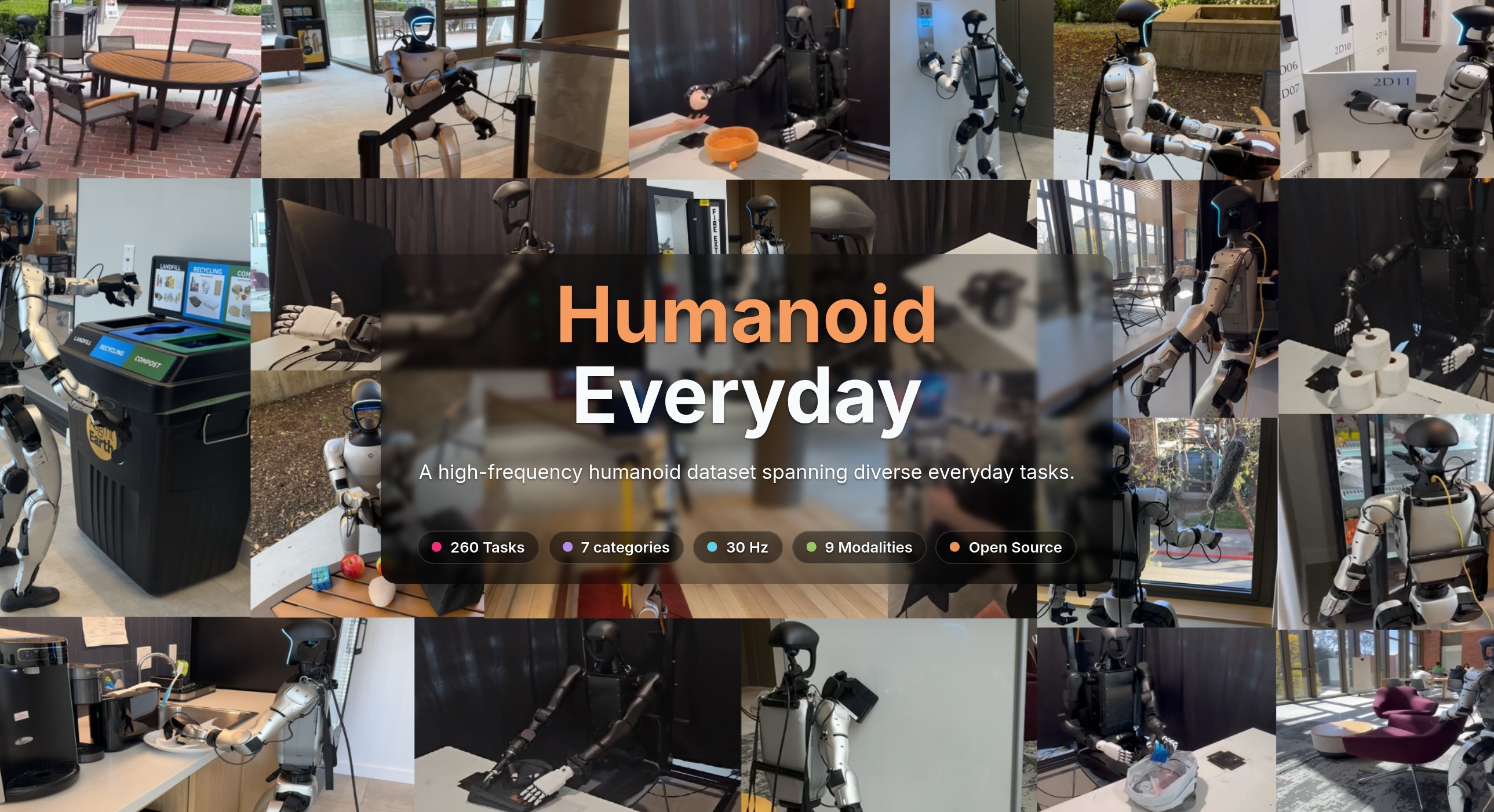}
  \captionof{figure}{\textbf{Humanoid Everyday}, a large-scale and diverse humanoid manipulation dataset characterized by extensive task varieties across a wide range of environments.}
  \label{fig:teaser}
  % \vspace{-2mm}
\end{strip}

\begin{abstract}
From loco-motion to dextrous manipulation, humanoid robots have made remarkable strides in demonstrating complex full-body capabilities. However, the majority of current robot learning datasets and benchmarks mainly focus on stationary robot arms, and the few existing humanoid datasets are either confined to fixed environments or limited in task diversity, often lacking human-humanoid interaction and lower-body locomotion. Moreover, there are a few standardized evaluation platforms for benchmarking learning-based policies on humanoid data. In this work, we present Humanoid Everyday, a large-scale and diverse humanoid manipulation dataset characterized by extensive task variety involving dextrous object manipulation, human-humanoid interaction, locomotion-integrated actions, and more. Leveraging a highly efficient human-supervised teleoperation pipeline, Humanoid Everyday aggregates high-quality multimodal sensory data—including RGB, depth, LiDAR, and tactile inputs—together with natural language annotations, comprising 10.3k trajectories and over 3 million frames of data across 260 tasks across 7 broad categories. In addition, we conduct an analysis of representative policy learning methods on our dataset, providing insights into their strengths and limitations across different task categories. For standardized evaluation, we introduce a cloud-based evaluation platform that allows researchers to seamlessly deploy their policies in our controlled setting and receive performance feedback. By releasing Humanoid Everyday along with our policy learning analysis and a standardized cloud-based evaluation platform, we intend to advance research in general-purpose humanoid manipulation and lay the groundwork for more capable and embodied robotic agents in real-world scenarios. Our dataset, data collection code, and cloud evaluation website are made publicly available on our project website: \url{https://humanoideveryday.github.io}
\end{abstract}

% \begin{IEEEkeywords}
% Humanoid robot, manipulation dataset, policy benchmarking, cloud-based evaluation
% \end{IEEEkeywords}

\begin{figure*}[t]
  \centering
  \includegraphics[width=1.0\textwidth]{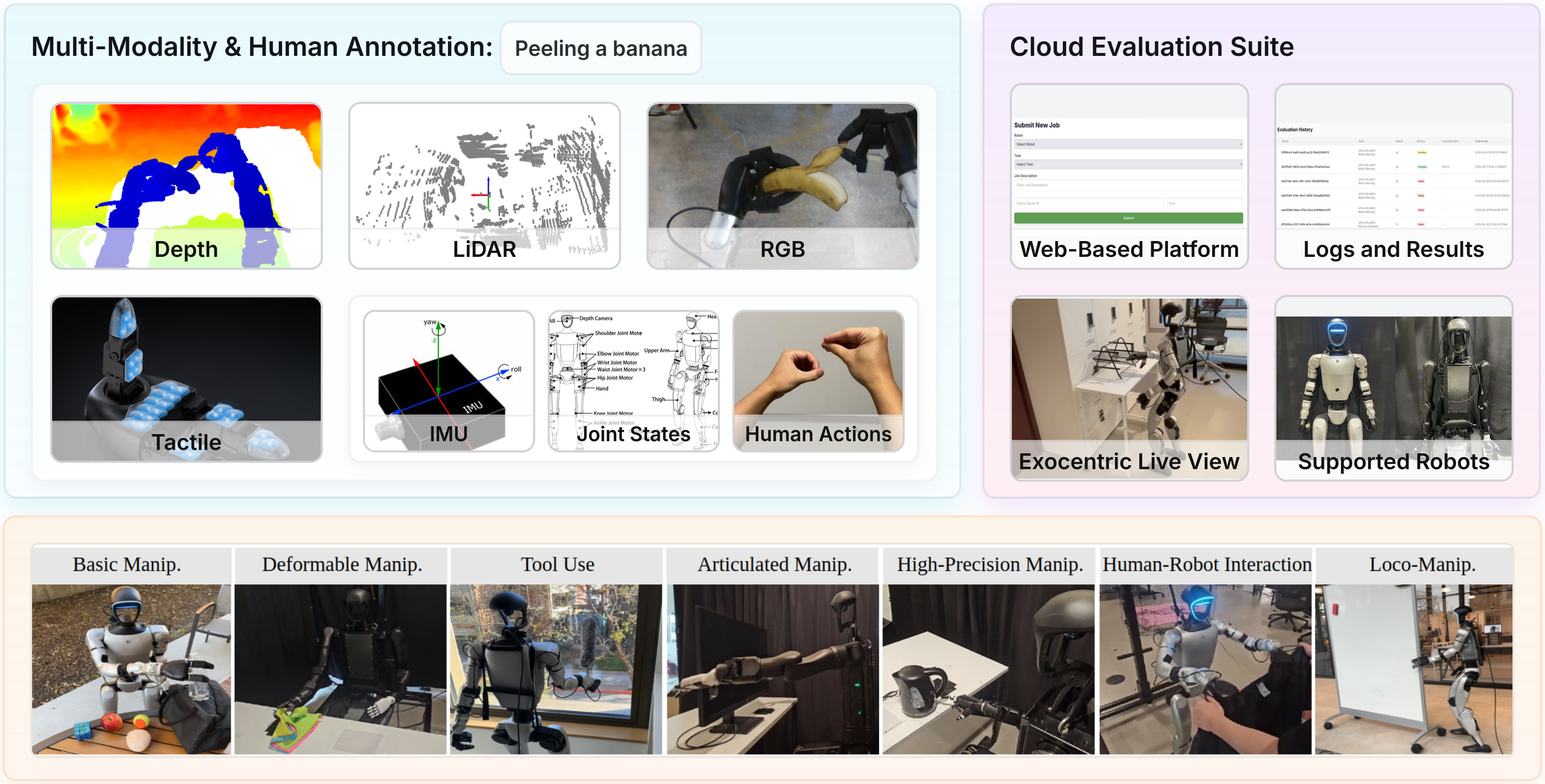}
  \caption{\textbf{Humanoid Everyday Overview.} Humanoid Everyday covers 7 distinct categories of humanoid manipulation tasks with rich multimodal information, and provides a cloud-based evaluation platform for standardized policy deployment.}
  \label{fig:block}
\end{figure*}
\section{Introduction}

% Recent years have witnessed rapid advancements in humanoid robotics, significantly narrowing the embodiment gap between humans and humanoid robots. From running and dancing to performing complex full-body movements~\cite{ji2024exbody2, cheng2024expressive, xue2025unified, mao2024learning}, modern humanoid robots continue to demonstrate their remarkable capabilities that could bring them closer to seamless human interaction. Among these abilities, manipulation stands out as particularly vital. The ability to physically interact with and manipulate objects in diverse environments is key to enabling humanoid robots to assist with real-world tasks, both in domestic and industrial settings. As robots become more integrated into daily routines, there is a growing demand for humanoid manipulation systems that are not only capable, but also robust and adaptable. \par

Recent progress in humanoid robotics has significantly reduced the embodiment gap, enabling robots to perform dynamic activities such as running, dancing, and complex full-body movements~\cite{ji2024exbody2, cheng2024expressive, xue2025unified, mao2024learning}. However, collecting humanoid manipulation datasets remains challenging. It requires operating in both indoor and outdoor environments, executing a wide range of tasks, and leveraging the humanoid form through bimanual coordination, full-body motion, and human-centric interactions. Existing datasets mainly target stationary arms or mobile platforms with simple grippers and wheeled bases~\cite{o2024open, fang2023rh20t, walke2023bridgedata, fourier2025actionnet, bu2025agibot} (see~\autoref{tab:dataset-comparison}); even egocentric efforts like Humanoid Policy~\cite{qiu2025-humanpolicy} emphasize repetitive tasks with limited locomotion. These limitations underscore the need for a diverse, interactive humanoid manipulation dataset that captures the full spectrum of human-like capabilities across varied environments and task complexities.\par

Beyond the scarcity of diverse and interactive datasets, there is also a notable lack of standardized evaluation practices for humanoid manipulation. Although recent deep learning methods have demonstrated improving performance in robotic manipulation~\cite{chi2023diffusion, ze20243d, zhao2023learning, kim2024openvla}, there remains a need for a unified evaluation framework for systematically comparing these approaches in the context of humanoid tasks. The absence of an evaluation standard makes it challenging to conduct fair and rigorous comparisons across policies, limiting our understanding of what truly drives effective humanoid manipulation in diverse scenarios. Together, these limitations lead to a ponderous question: what should an effective humanoid manipulation dataset look like to bridge the gaps in diversity, embodiment, and evaluation in the development of more intelligent robotic agents?\par

In this work, we introduce the Humanoid Everyday Dataset, a large-scale collection of 260 tasks across 7 categories (see \autoref{fig:block}) that captures full-body locomotion, dexterous manipulation, and rich human-humanoid interactions in diverse environments. Unlike prior datasets limited in scope or simple settings~\cite{qiu2025-humanpolicy, fourier2025actionnet, bu2025agibot}, we accumulate more comprehensive and human-like tasks across various environments, retaining human supervision for data accuracy. Our additional technical re-engineering of the Unitree official teleoperation script enables sub-millisecond data synchronization across different modalities. At 30Hz, our pipeline captures high-resolution sensor and action data from every humanoid episode. We record each task with egocentric RGB video, depth information, LiDAR scans, tactile and Inertia Measurement Unit (IMU) information, joint poses, joint actions, and natural language annotations.\par 

Beyond data collection, we analyzed representative policy learning methods for Humanoid Everyday, highlighting strengths and limitations across different task categories. These analyses provide initial insights into the challenges of embodied humanoid manipulation. To further support the community, we introduce a cloud-based evaluation platform that allows researchers to upload their policies, execute them within our standardized real-world environment, and receive detailed performance feedback. Unlike existing cloud evaluation systems that focus on robotic arms~\cite{zhou2025autoeval} or simulated agents~\cite{li2024evaluating}, ours is the first platform designed for humanoid, which aims to lower the barrier to fair comparison and fosters collaborative progress in humanoid robotics. Together, Humanoid Everyday and our evaluation system offer a valuable foundation for advancing research on general-purpose humanoid manipulation for developing more robust, capable, and embodied agents in real-world scenarios.\par

Our contributions are threefold: (1) a large-scale multimodal humanoid manipulation dataset collected in diverse real-world scenarios with an optimized teleoperation pipeline; (2) an analysis of representative policy learning methods on Humanoid Everyday, highlighting their strengths and limitations across task categories; and (3) a cloud-based evaluation platform that enables standardized, reproducible, and collaborative research in humanoid manipulation.

\begin{strip}
\centering

\begin{adjustbox}{width=\textwidth}
\begin{tabular}{
    @{}l     % Dataset
    r        % # Tasks.
    r        % # Skills
    c        % Lang. Annotation
    c        % Humanoid Robot       
    c        % Hand Type
    c        % Bipedal loco-manipulation
    c
    c@{}     % Collection
}
\toprule
Dataset & 
\# Trajectories & 
\# Skills & 
\begin{tabular}[c]{@{}c@{}}Lidar\\Sensing\end{tabular} & 
\begin{tabular}[c]{@{}c@{}}Humanoid\\Robot\end{tabular} &  
\begin{tabular}[c]{@{}c@{}}Dexterous\\Hand\end{tabular} &  
\begin{tabular}[c]{@{}c@{}}Bipedal\\Loco-Manipulation\end{tabular} & 
\begin{tabular}[c]{@{}c@{}}Human-Robot\\Interaction\end{tabular} &
\begin{tabular}[c]{@{}c@{}}Cloud\\Evaluation\end{tabular} \\
\midrule
MIME~\cite{sharma2018multiple}     & 8.3k   & 12 & \xmark & \xmark & \xmark & \xmark & \xmark & \xmark \\
RoboTurk~\cite{mandlekar2018roboturk}     & 2.1k   & 2 & \xmark & \xmark & \xmark & \xmark & \xmark & \xmark \\
RoboNet~\cite{dasari2019robonet}         & 162k  & N/A  & \xmark & \xmark & \xmark & \xmark & \xmark & \xmark \\
BridgeData~\cite{ebert2021bridge}     & 7.2k  & 4   & \xmark & \xmark & \xmark & \xmark & \xmark & \xmark \\
RH20T~\cite{fang2023rh20t}      & 13k   & 33   & \xmark & \xmark & \xmark & \xmark & \xmark & \xmark \\
DROID~\cite{khazatsky2024droid}   & 76k  & 86   & \xmark & \xmark & \xmark & \xmark & \xmark & \xmark \\
Open X-Embodiment~\cite{o2024open}         & 1.4M  & 217   & \xmark & \xmark & \xmark & \xmark & \xmark & \xmark \\
Fourier ActionNet~\cite{fourier2025actionnet}      & 13k  & 16  & \xmark & \cmark & \cmark & \xmark & \xmark & \xmark \\
Agibot World~\cite{bu2025agibot}     & 1M  & 87   & \xmark & \xmark* & \xmark$^\dag$ & \xmark & \cmark & \xmark \\
\midrule
\textbf{Humanoid Everyday} & 
10.3k & 
221 & 
\cmark & \cmark & \cmark & \cmark & \cmark & \cmark \\
\bottomrule
\end{tabular}
\end{adjustbox}

\vspace{1mm}
\raggedright{\footnotesize $^*$~Wheeled base instead of bipedal legs, not considered a humanoid robot here.
$^\dag$~Mostly grippers and only a few dexterous hands.}

\captionsetup{justification=centering}
\captionof{table}{Comparison of Robotic Manipulation Datasets}
\label{tab:dataset-comparison}

\end{strip}

\section{Related Work}

\subsection{Robotic Manipulation Datasets}
Robot manipulation datasets have played an important role in enabling data-driven robot policy learning in recent years. Some datasets focus on a single pattern of behavior, such as pushing~\cite{dasari2019robonet, finn2017deep}, grasping~\cite{depierre2018jacquard, zhao2025robot}, and pouring~\cite{lin2023pourit, burns2022look}. Some other large-scale multi-task datasets have also been proposed to improve generalization across diverse manipulation scenarios and environmental conditions, but still focus on robotic arm platforms~\cite{o2024open, walke2023bridgedata, bousmalis2023robocat}. More recently, the rise of humanoid robots has led to the creation of humanoid-specific datasets. ~\cite{wagener2022mocapact, qiu2025-humanpolicy, fourier2025actionnet} leverage motion capture data or human teleoperation to train humanoid agents in either simulated or constrained tabletop settings, focusing on simplified manipulation tasks. ~\cite{ze2024generalizable, bu2025agibot} collect real-world humanoid robotic data, demonstrating diverse manipulation skills using upper-body teleoperation.
Despite these advances, most existing datasets focus on robotic-arm manipulation or humanoids with limited capabilities, often lacking diverse tasks, complex environments, and full-body functions. In contrast, Humanoid Everyday captures a broad spectrum of full-body humanoid activities across various indoor and outdoor settings, offering a comprehensive resource for general-purpose humanoid policy development.\par

% \begin{figure*}[t]
%   \centering

%   \begin{subfigure}[t]{0.32\textwidth} 
%     \centering
%     \includegraphics[width=\linewidth]{images/pp.png}
%     \label{fig:a}
%   \end{subfigure}
%   \hspace{1em}
%   \begin{subfigure}[t]{0.3\textwidth}
%     \centering
%     \fbox{\rule{0pt}{4cm} \rule{4cm}{0pt}}
%     \label{fig:b}
%   \end{subfigure}
%   \hspace{1em}
%   \begin{subfigure}[t]{0.3\textwidth}
%     \centering
%     \includegraphics[width=\linewidth]{images/platform.jpg}
%     \label{fig:c}
%   \end{subfigure}

%   \caption{\textbf{Comprehensive Pipeline for Data Collection and Policy Evaluation.} \textbf{(a)} Our teleoperation system enables multi-modal data acquisition, incluing Ego-centric images, depth, LiDAR and tactile information. \textbf{(b)} We build simulation environments for tasks in Humanoid Everyday, aiming at early-stage policy verification. \textbf{(c)} Our online evaluation platform allows users to upload their policies and test on our real-world robot settings for fair benchmarking.}
%   \label{1em}
% \end{figure*}

\subsection{Humanoid Robot Learning}
Recent works have explored diverse learning-based policies on humanoid robots. Many of them leverage reinforcement learning and sim-to-real transfer to achieve whole‐body coordination for balance and locomotion~\cite{li2023robust, radosavovic2024humanoid, seo2023deep, tang2024humanmimic, he2025asap, ji2024exbody2}, some others also incorporate human motion datasets in the RL training pipeline to ease sim-to-real transfer~\cite{cheng2024expressive, he2024omnih2o}. Beyond whole-body control, learning humanoid manipulation has also become an active research direction. Many approaches begin by collecting demonstration data through teleoperation using VR applications~\cite{cheng2024open, fu2024humanplus, he2024learning, lu2024mobile}. Imitation learning and Vision-Language-Action (VLA) models are then utilized to train corresponding policies, allowing robots to replicate demonstrated behaviors with high fidelity~\cite{mees2022calvin, mees2022matters, cheng2024expressive, fu2024humanplus, he2024learning, peng2021amp, fu2024mobile, li2025amo}. However, current approaches are inherently limited by the diversity of collected demonstrations, which mostly concentrate on a narrow set of tasks or environments. Consequently, few works have systematically evaluated humanoid policies across diverse task categories. Humanoid Everyday offers the diversity and scale needed to train more robust and generalizable manipulation policies, and our analysis of representative imitation learning approaches provides insights into their strengths and limitations, thereby facilitating more efficient learning and execution of complex behaviors.\par

\subsection{Robot Policy Evaluation}
Beyond the availability of humanoid datasets and training methods, fair evaluation of robotic policies is critical for measuring progress and ensuring reproducibility across diverse systems and environments. 
% Physics simulation platforms~\cite{juliani2018unity, coumans2015bullet, authors2024genesis, xiang2020sapien} have been proposed to emulate real-world physics modeling of collisions, contacts, and complex dynamics for robotic policy development and evaluation. 
~\cite{lee2021ikea, mees2022calvin, yu2020meta, wang2025roboeval} develop benchmarking suites and evaluation protocols that assess robotic performance in simulation systems under controlled conditions. In addition to simulation-based evaluation, directly assessing policies in real-world settings is also a common practice. ~\cite{yang2019replab, heo2023furniturebench, luo2025fmb} have proposed simple and reproducible real-world robotic setups to enable consistent policy inference and evaluation under similar environments. AutoEval~\cite{zhou2025autoeval} allows users to upload and deploy different learned policies under the same standardized setting for autonomous evaluation. However, these evaluation frameworks are largely confined to robotic arm platforms and do not extend to humanoid robots with their unique challenges. To fill this gap, Humanoid Everyday introduces a cloud-based evaluation platform tailored for humanoid robots. Our platform enables deployment on a standardized real-world humanoid system, providing consistent inference and evaluation for benchmarking manipulation policies across diverse tasks and users.

\section{Humanoid Everyday Dataset}
To support learning and evaluation of humanoid manipulation across various real-world tasks, we introduce the Humanoid Everyday Dataset, a large-scale, high-quality dataset collected with full-body humanoid robots, spanning diverse tasks including human-humanoid interaction and loco-manipulation. In Section~\ref{subsec:env_setup}, we describe the hardware setup, including the humanoid platforms and the operation interface used for data collection. Section~\ref{subsec:pipeline} outlines our data collection pipeline and describes how our method improves the overall efficiency. Finally, we present the composition and structure of the dataset in Section~\ref{subsec:dataset}.

\label{headings}

% \begin{figure}[H]
%   \centering
%  %\includegraphics[width=0.7\linewidth]{images/pp.png}
%   \includegraphics[width=\linewidth]{images/datacollection.pdf}
%   \caption{\textbf{Data Collection.} Tasks in Humanoid Everyday are captured using a rich set of sensory modalities.}
%   \label{fig:collection}
% \end{figure}

\subsection{Environment setup} \label{subsec:env_setup}

\paragraph{Hardware}
We gather data with two Unitree humanoid robots: the 29-degrees of freedom (DoF) \textbf{G1} with 7-DoF three-fingered dexterous hands (Dex3-1) and the 27-DoF \textbf{H1} with 6-DoF INSPIRE hands. Both the H1 and G1 humanoid robots are equipped with Intel RealSense RGB-D cameras and a Livox LiDAR system. In addition, the G1's Dex3-1 hands possess tactile sensors, which further enhances the multimodality of our dataset. 
% \jiageng{DoFs of the G1 and H1 body?}

\paragraph{Teleoperation interface}
The operator wears an Apple Vision Pro to capture wrist and finger keypoints using its cameras on the bottom. The finger motions are mapped onto the robot’s dexterous hands via the dex-retargeting system~\cite{qin2023anyteleop} for the robot hands to complete basic manipulations. The wrist poses are converted into arm joint commands through a Pinocchio-based inverse kinematics algorithm~\cite{carpentierpinocchio}, enabling full upper-body teleoperation.

\subsection{Efficient and Scalable Data Collection} \label{subsec:pipeline}
We propose a multi-processing teleoperation pipeline, re-engineered upon the official Unitree teleoperation library and significantly improved for large-scale, high-quality humanoid data collection. Our intuitive and robust design enables low-latency teleoperation with high-frequency control while ensuring synchronized, high-quality data streams.\par

\begin{figure}[H]
  \centering
  \begin{subfigure}{\linewidth}
    \centering
    \includegraphics[width=0.85\linewidth]{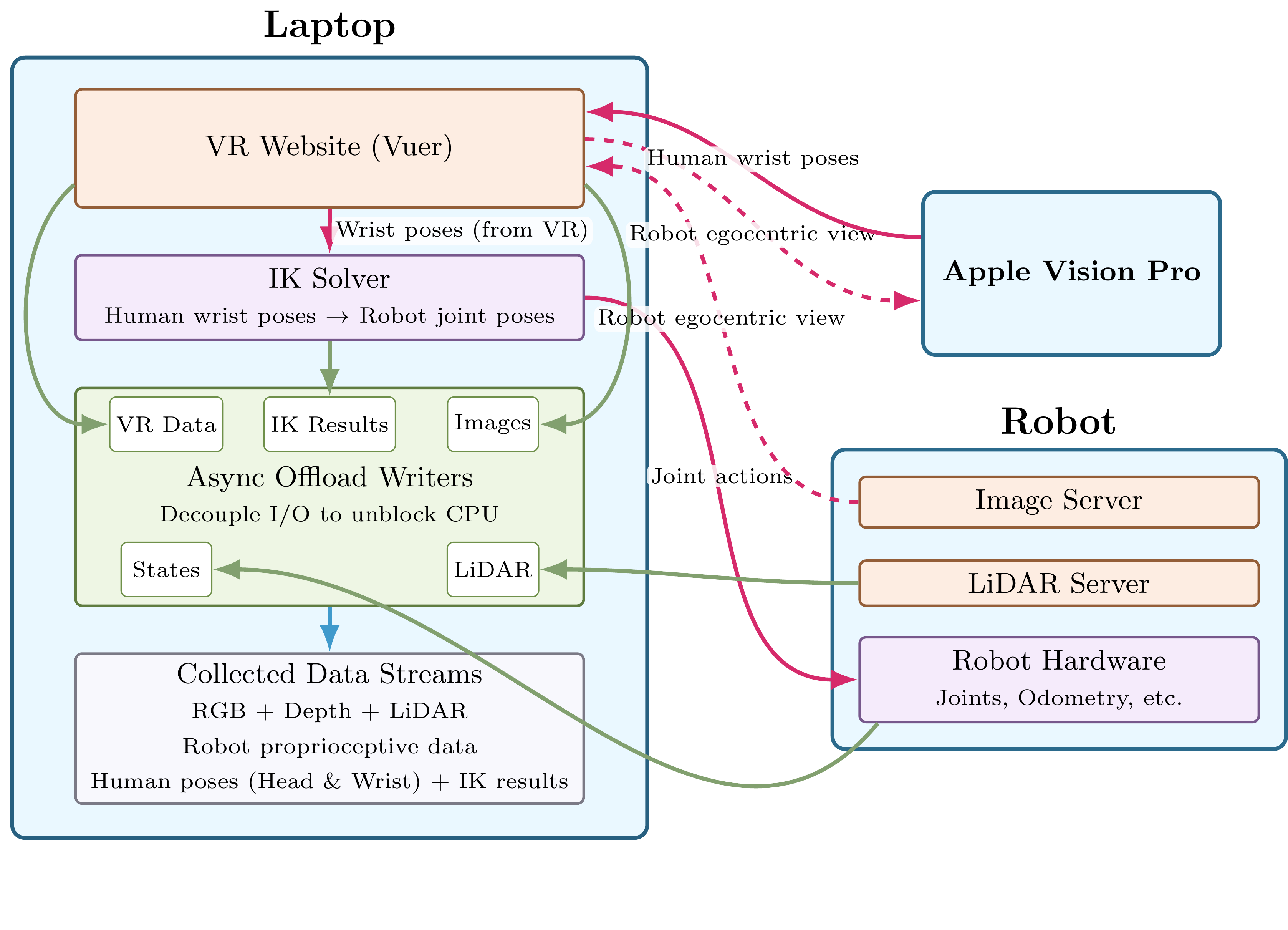}
    \caption{}
    \label{fig:collection-a}
  \end{subfigure}
  
  \begin{subfigure}{\linewidth}
    \centering
    \includegraphics[width=0.85\linewidth]{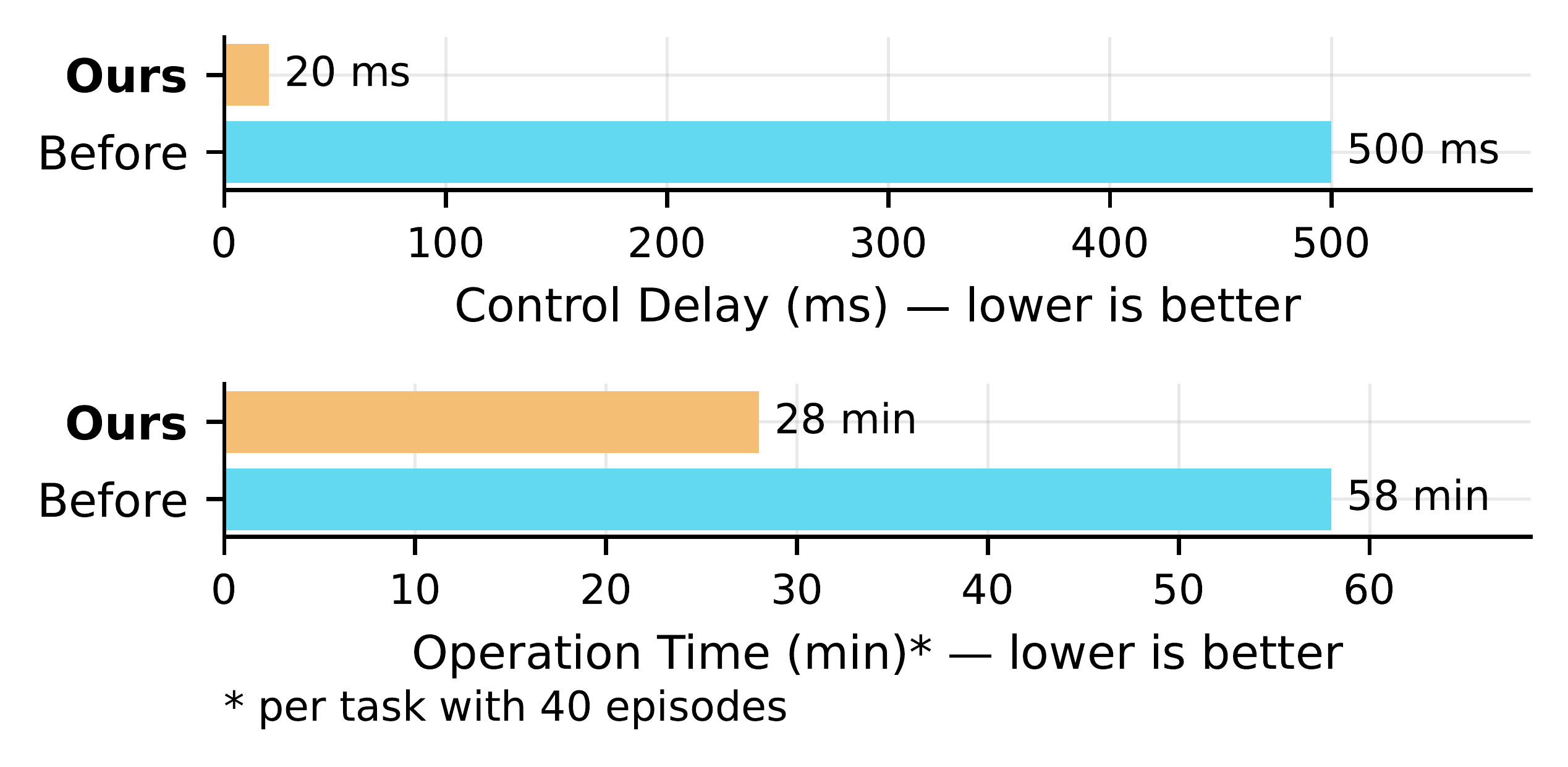}
    \caption{}
    \label{fig:collection-b}
  \end{subfigure}

  \caption{\textbf{Data Collection Pipeline.} 
  (a) We separate data streaming, data writing, robot control, and IK computation into distinct processes and threads to ensure reliable and efficient data collection.  
  (b) Our pipeline substantially reduces control delay and enhances data collection efficiency.}
  \label{fig:collection}
\end{figure}

As opposed to the blocking and synchronous design in the Unitree official teleoperation script, our data collection pipeline uses multi-processing and asynchronous IO reading and writing to ensure high-frequency teleoperation and high-quality data collection. Specifically, as shown in~\autoref{fig:collection-a}, we decouple IO data from inverse kinematics (IK) computation and robot joint control in separate processes, with shared memory buffers facilitating fast and low-latency inter-process communication. This design allocates more computational resources to the IK solver, enabling smoother and higher-frequency teleoperation control. In addition, data recording and sensor processing are handled asynchronously in parallel threads within main pipeline process, ensuring nonblocking and temporally aligned data collection.\par

Additionally, we provide a simple data-collection interface that hides system complexity, stream the robot’s binocular IR feeds to the VR headset for better situational awareness, and support multiple recording sessions per run without restarting the entire program.

All data collection was performed on a laptop equipped with an 11th Gen Intel i7 CPU. As shown in~\autoref{fig:collection-b}, our pipeline halves data collection time compared to the official Unitree teleoperation system, while the control delay decreases from 500 ms to 2 ms. These improvements highlight the efficiency and scalability of our pipeline, enabling rapid and high-quality data acquisition for complex humanoid tasks.

% \begin{figure}[H]
%   \centering
%   \includegraphics[width=1.0\linewidth]{images/operation.pdf}
%   \caption{Comparison of data collection efficiency and control performance.}
%   \label{fig:efficiency}
% \end{figure}

\subsection{A Diverse Collection of Everyday Humanoid Tasks} \label{subsec:dataset}
\begin{figure}[H]
  \centering
  \includegraphics[width=1.0\linewidth]{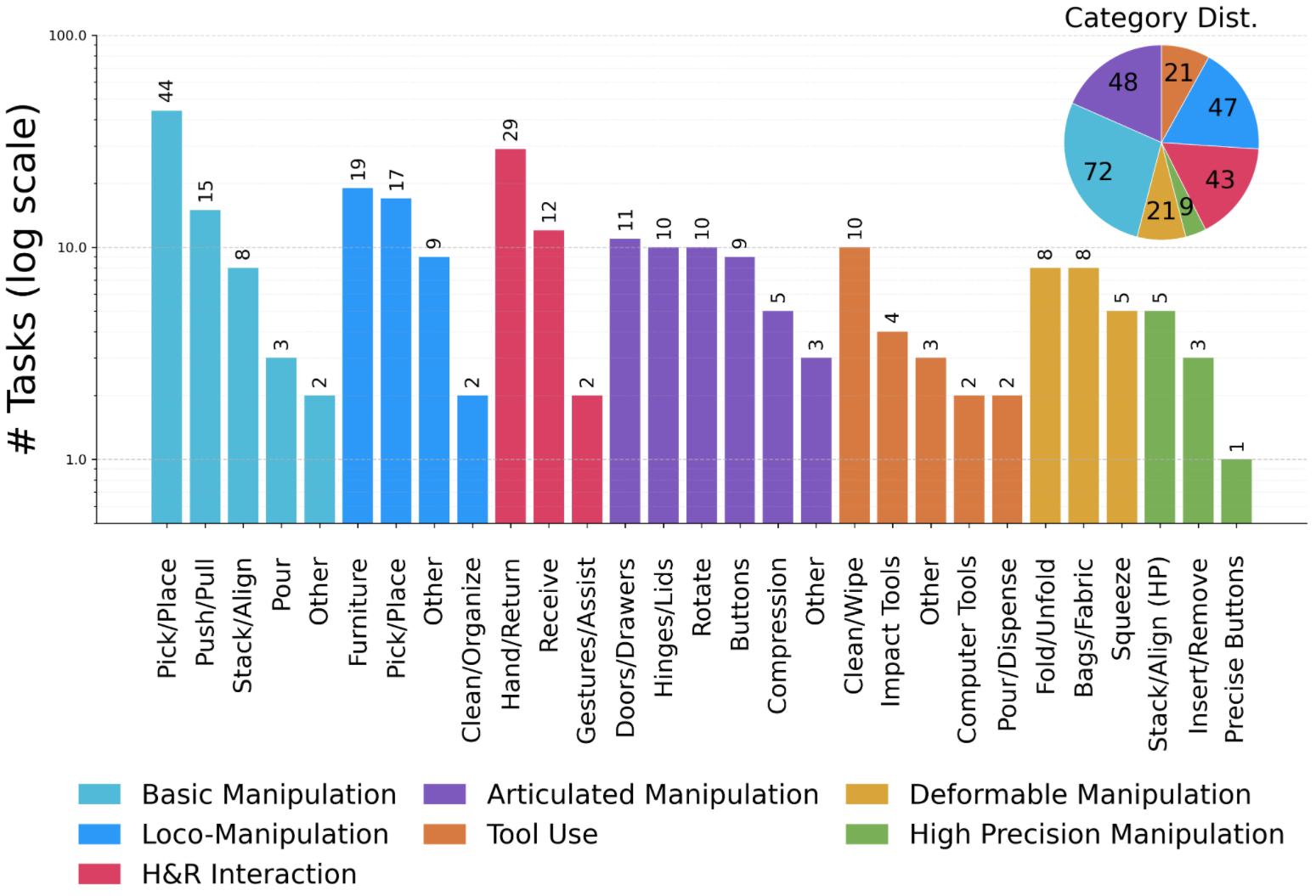}
  \caption{\textbf{Data Distribution.} Distribution of tasks and skill categories in the Humanoid Everyday Dataset.}
  \label{fig:distribution}
\end{figure}

Humanoid Everyday is a large-scale, diverse collection of humanoid manipulation tasks, comprising seven main distinct categories including Basic Manipulation (basic object pick-and-place manipulation), Deformable Manipulation (interacting with cloths or other deformable objects), Articulated Manipulation (operating hinged or jointed structures), Tool Use (utilizing external objects to achieve goals), High-Precision Manipulation (performing difficult tasks requiring high accuracy), Human-Robot Interaction (engaging in cooperative actions with humans), and Loco-Manipulation (combining locomotion and manipulation) (see~\autoref{fig:distribution}. Our tasks are performed in indoor and outdoor environments, involving complex interactions with surrounding objects and dynamic settings. Additionally, a portion of these tasks require lower-body locomotion, adding further diversity to the dataset. This variety in environmental contexts enriches the collected data and supports the development of generalizable policies capable of adapting to diverse real-world scenarios.\par

% Humanoid Everyday includes 260 unique tasks in total, each recorded with around 40 episodes to ensure sufficient data for effective training. Besides, every task in the dataset is captured using a rich set of sensory modalities, providing a comprehensive understanding of the humanoid’s interactions. Specifically, each episode includes RGB video for visual perception; depth information; LiDAR scans for 3D scene understanding; tactile data for contact and force feedback; and natural language descriptions that offer semantic context for the task. These multimodal data provide richer training information, enabling the learning of more adaptable and context-aware humanoid policies.\par

Humanoid Everyday comprises 260 unique tasks, each with ~40 episodes to provide sufficient training data. We record each task with a rich set of sensory modalities, offering a comprehensive view of humanoid interactions. Each episode includes RGB video, depth maps, LiDAR, tactile feedback, and natural language task descriptions. This multimodal design enables richer training trajectories, supporting the development of more adaptable and context-aware humanoid policies.\par

\section{Evaluating Humanoid Manipulation Policies}
\begin{figure}[H]
  \centering
  \includegraphics[width=1.0\linewidth]{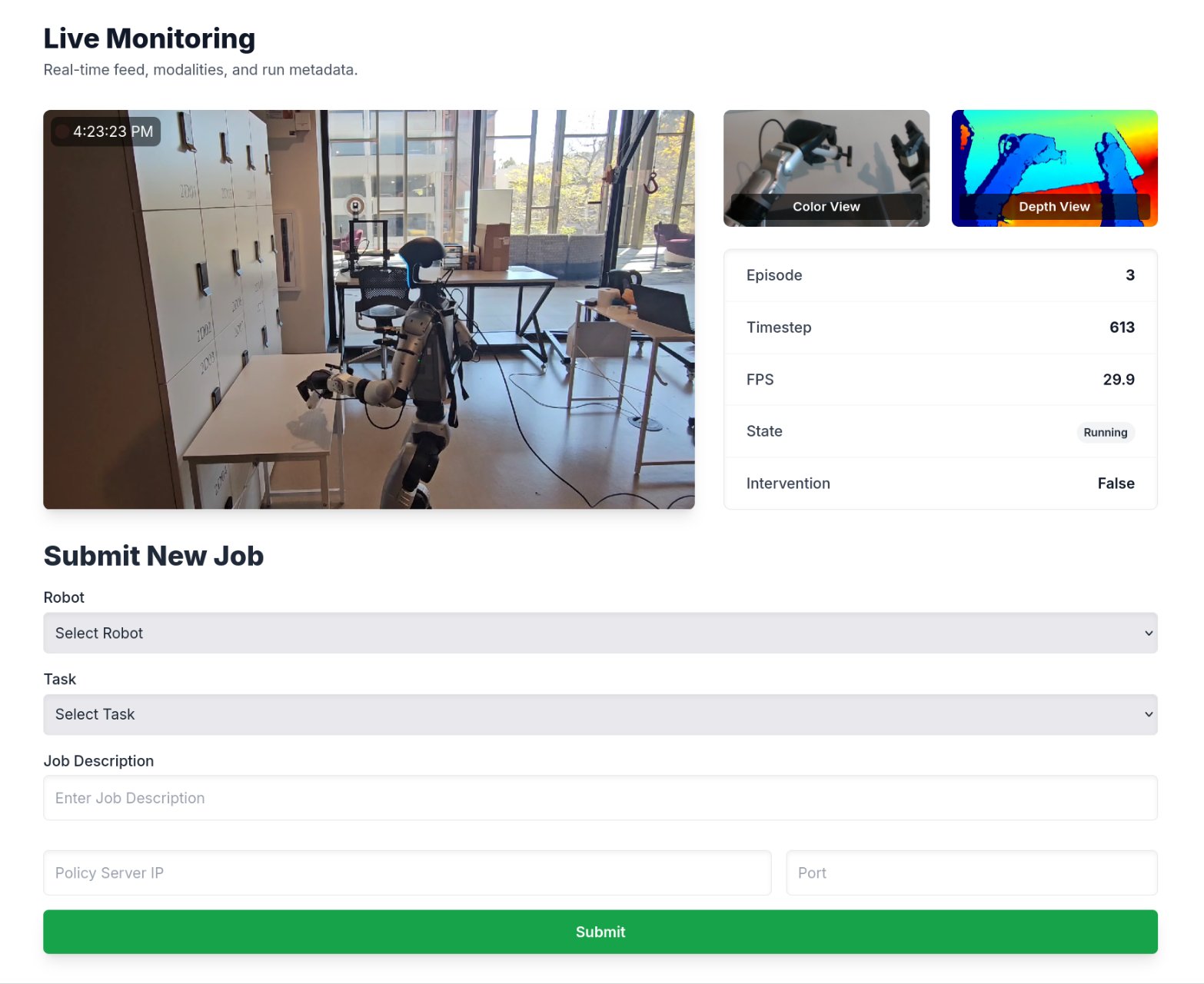}
  \caption{\textbf{Evaluation platform.} Humanoid Everyday introduces a cloud-based evaluation platform on real-world humanoid setups }
  \label{fig:eval}
\end{figure}

To enable systematic and reproducible evaluation of humanoid manipulation policies, we introduce a cloud-based evaluation platform tailored for humanoid robots, as shown in~\autoref{fig:eval}.

Our cloud-based evaluation platform enables remote policy deployment on locally hosted humanoid robots, addressing the hardware access bottleneck in real-world policy inference. By allowing researchers to test their policies on actual humanoid robots without owning them, our platform significantly lowers the barrier to humanoid manipulation learning research and standardizes the evaluation process across different methods and users.\par

We reconstruct a subset of task environments from the Humanoid Everyday Dataset to support realistic and standardized evaluations. Researchers wishing to evaluate their trained policies over the Humanoid Everyday Dataset can connect to the platform by simply specifying their policy server IP and port, after which our system streams real-time visual inputs (RGB images and Depth information) and robot state information from the G1 or H1 robot to the client. Users can then run inference using their own policies locally and send the resulting action commands back to our server. These commands are executed in real time on a physical humanoid robot. In addition, we stream both egocentric RGB and depth images from the robot and third-person camera views to a web interface, allowing users to monitor the task execution remotely.\par

\begin{figure}[H]
  \centering
  \includegraphics[width=1.0\linewidth]{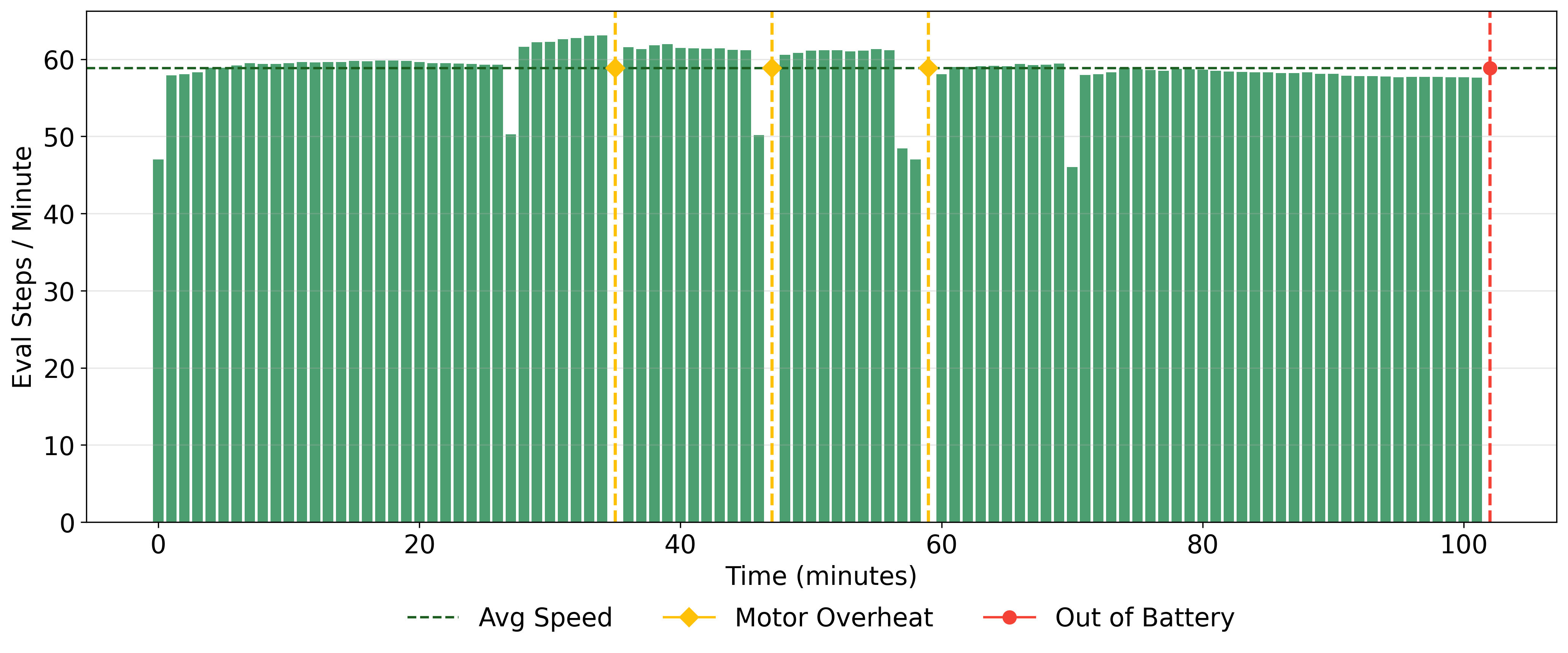}
  \caption{Evaluation Steps per Minute with Human Interventions. Our evaluation runs continuously for over 100 minutes before running out of battery, while only three human interventions were required due to motor overheating. The system maintained consistently high evaluation efficiency throughout the rest of the process.}
  \label{fig:intervention}
\end{figure}

Our evaluation platform framework enables highly efficient deployment of humanoid policies. As shown in~\autoref{fig:intervention}, our online evaluation system runs continuously for over 100 minutes before battery depletion. Only three human interventions were required due to motor overheating, and the system otherwise maintained high evaluation efficiency throughout the rest of the process. This cloud-based system enables reproducible and hardware-agnostic evaluation of humanoid manipulation policies on our Humanoid Everyday dataset. We believe it could provide a practical solution for researchers without direct access to humanoid robots, and could establish a common testbed for fair benchmarking and collaborative development.\par

\begin{table*}[!b]
  \centering
  \setlength{\tabcolsep}{8pt}
  \renewcommand{\arraystretch}{1.15}
  \begin{tabularx}{\textwidth}{X X c c c c c c c}
    \toprule
    \textbf{Task Category} & \textbf{Task} & \textbf{DP} & \textbf{DP3} & \textbf{ACT} & \textbf{OpenVLA} & \textbf{$\pi_{0}$-FAST} & \textbf{$\pi_{0.5}$} & \textbf{GR00T N1.5} \\
    \midrule
    Articulate & Rotate chair                      & 100\% & 90\% & 100\% & 70\% & 100\% & 100\% & 100\% \\
    Tool Use & Use eraser to wipe the desk           & 0\%    & 70\%   & 0\%    & 30\%    & 40\%    & 40\%   & 0\%\\
    Basic & Put dumpling toy into plate      & 30\% & 20\%   & 70\% & 30\% & 60\% & 30\% & 80\% \\
    Deformable & Fold towel on the desk           & 0\%    & 20\%   & 0\%    & 40\%    & 20\%    & 40\%  & 50\%\\
    HRI & Hand over dumpling toy             & 40\%    & 40\%   & 70\%    & 60\%    & 30\%    & 40\%  & 100\%\\
    Loco-Manip. & Walk to grab door handle             & 30\%   & 0\%   & 0\%    & 30\%    & 10\%    & 0\%  & 30\%\\
    High Precision & Insert rose into vase           & 0\%    & 0\%   & 0\%    & 10\%    & 0\%    & 0\%  & 0\%\\
    \midrule
    \textbf{Average} &  & 29\% & 34\% & 34\% & 39\% & 37\% & 36\% & 51\% \\
    \bottomrule
  \end{tabularx}
  \caption{Success rates of imitation learning methods on the Humanoid Everyday Dataset.}.
  \label{tab:experiments}
\end{table*}

\begin{figure*}[t]
  \centering
  \includegraphics[width=1.0\textwidth]{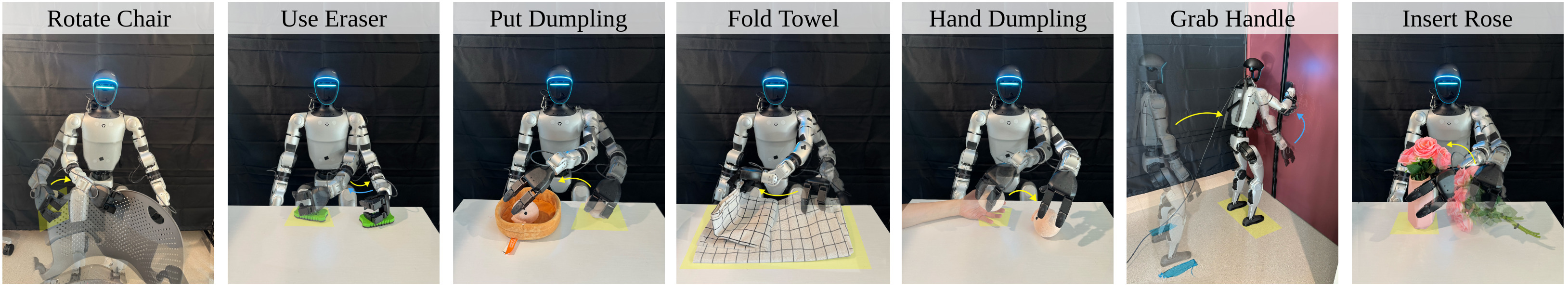}
  \caption{\textbf{Experiment Setup}. Representative inference tasks from the seven categories. The yellow region denotes the task execution area with slight variations, and arrows show the robot’s arm trajectories. For each task, we conduct 10 trials under seven different policies.}
  \label{fig:experiments}
\end{figure*}

\section{Experiments}
\subsection{Performance of Imitation Learning Policies}
To better understand how existing imitation learning approaches perform in the context of humanoid manipulation, we evaluate a variety of policies on our Humanoid Everyday dataset, including Diffusion Policy (DP)~\cite{chi2023diffusion}, 3D Diffusion Policy (DP3)~\cite{ze20243d}, Action Chunking with Transformers (ACT)~\cite{zhao2023learning}, OpenVLA~\cite{kim2024openvla}, $\pi_{0}$-FAST~\cite{pertsch2025fast}, $\pi_{0.5}$~\cite{intelligence2025pi05visionlanguageactionmodelopenworld}, and GR00T N1.5~\cite{bjorck2025gr00tn1p5}. 
We train these policies on our 30Hz humanoid data until convergence.
For VLA-based policies, we adopt a two-stage finetuning strategy: we first fine-tune on the full Humanoid Everyday dataset and then further adapt each model to task-specific data from individual categories.\par

We conduct experiments across all seven task categories in Humanoid Everyday. For detailed per task experiment setup, see~\autoref{fig:experiments}. We present the results of these policies in~\autoref{tab:experiments}. where each policy is evaluated over 10 trials on seven tasks.

Overall, all the end-to-end imitation policies struggle in humanoid manipulation tasks due to the high-dimensional action space in our dataset (which is 28 DoFs in total). Among them, DP3 outperforms DP in most cases, and we hypothesize that the use of 3D point cloud observations provides stronger robustness to environmental variations. However, in Loco-Manipulation tasks where the robot must move and the point cloud undergoes large frame-to-frame changes, 3D-based inputs become less reliable than RGB images, leading to failure cases. ACT, on the other hand, performs poorly overall, as it does not effectively incorporate visual feedback and tends to overfit to demonstrated trajectories, mechanically replaying them without adapting to scene variations.\par

In contrast, large VLA models demonstrate more consistent and stable performance on humanoid manipulation, benefiting from pretraining priors that improve generalization, particularly on Deformable Manipulation and Loco-Manipulation tasks that require both precision and robustness. OpenVLA does not compress the action space and thus when trained on high-frequency 30 Hz data, it often fails to produce meaningful motions. Downsampling to 2 Hz alleviates this issue, but the resulting behaviors appear less smooth. $\pi_{0}$-FAST employs a DCT-based tokenizer for action compression, but the high-dimensional humanoid actions are not well represented, leading to excessive token numbers, incorrect output tokens, and decoding errors that sometimes cause the robot to halt during inference. On the other hand, although $\pi_{0.5}$ produces smoother humanoid actions compared to the previous two models, it still tends to overfit to trajectories, ignoring visual feedback. In particular, GR00T N1.5 achieves the strongest performance overall, largely due to its extensive pretraining across multiple large-scale humanoid datasets, which provides a powerful prior well-suited for our diverse manipulation tasks.\par

Despite these differences, all imitation learning policies struggle significantly on the most challenging categories, such as Loco-Manipulation and High-Precision Manipulation tasks. In the "Insert rose into vase" task, nearly all policies achieve a 0\% success rate: while many can lift the rose, they consistently fail to insert its thin stem into the vase, suggesting that current models lack fine-grained visuospatial perception. This indicates that imitation learning on humanoid platforms still faces significant challenges and opportunities for progress.

\subsection{Pretraining on Humanoid Everyday as a Prior}
To validate whether Humanoid Everyday can serve as an effective pretraining prior for large Vision-Language-Action (VLA) models, we design an ablation study comparing two training strategies: (i) the two-stage finetuning pipeline introduced in the previous section, and (ii) direct task-specific finetuning without pretraining on Humanoid Everyday. We select the Human–Robot Interaction task of handing over a dumpling toy, which represents a moderately challenging manipulation scenario. This task requires both coordination and robustness to interaction dynamics, yet is tractable enough for all VLA models to achieve a non-trivial level of success.

\begin{figure}[H]
  \centering
  \includegraphics[width=0.9\linewidth]{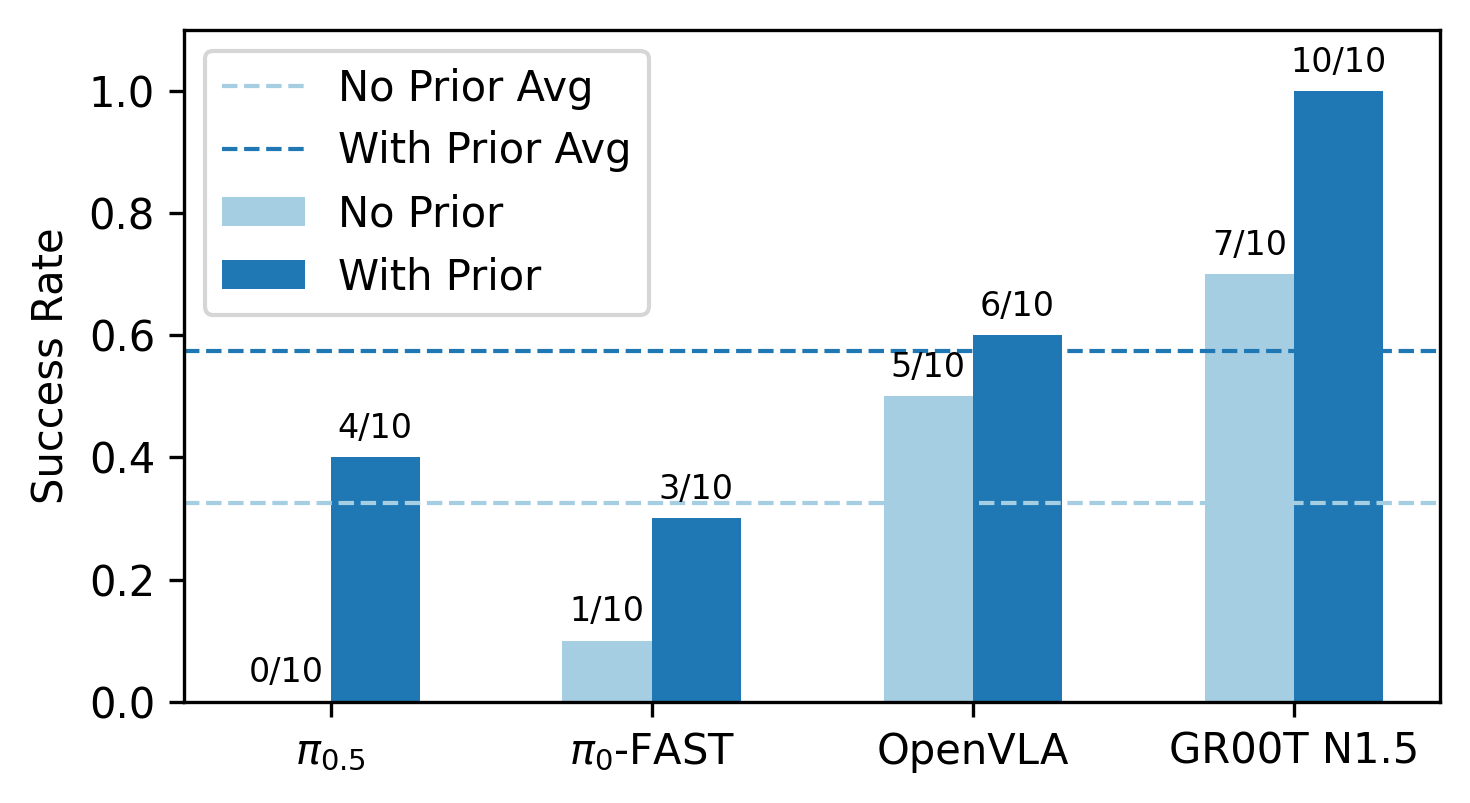}
  \caption{\textbf{Ablation on Humanoid Everyday Pretraining.} Performance comparison between direct task-specific finetuning and two-stage finetuning with Humanoid Everyday.}
  \label{fig:ablation}
\end{figure}

As shown in~\autoref{fig:ablation}, finetuning on Humanoid Everyday prior to task-specific adaptation consistently improves the performance of VLA models. This suggests that exposure to diverse humanoid behaviors provides a useful prior that facilitates learning downstream manipulation tasks, indicating that large-scale and diverse humanoid data can improve robustness and stability in humanoid manipulation.

% \begin{table}[!htbp]
%   \centering
%   \setlength{\tabcolsep}{8pt}
%   \renewcommand{\arraystretch}{1.15}
%   \begin{tabularx}{\columnwidth}{X c c}
%     \toprule
%     \textbf{Model} & \textbf{No Prior} & \textbf{With Prior} \\
%     \midrule
%     $\pi_{0.5}$     & 0/10 & 4/10 \\
%     $\pi_{0}$-FAST  & 1/10 & 3/10 \\
%     OpenVLA         & 5/10 & 6/10 \\
%     GR00T N1.5      & 7/10 & 10/10 \\
%     \midrule
%     \textbf{Average} & 0.33 & 0.58 \\
%     \bottomrule
%   \end{tabularx}
%   \caption{Effect of using the full Humanoid Everyday dataset as a prior for finetuning across models.}
%   \label{tab:finetune_prior}
% \end{table}

\section{Discussion and Conclusion}
We introduce Humanoid Everyday, a large-scale, diverse humanoid manipulation dataset encompassing full-body locomotion, dexterous manipulation, and rich human-humanoid interactions in various everyday settings. Leveraging this dataset, we conduct policy learning analyses that highlight both the strengths and shortcomings of current methods in humanoid manipulation. Additionally, we provide a cloud-based evaluation platform that enables policies to be deployed directly on our humanoid robot settings, promoting reproducibility and collaborative progress within the global humanoid robotics community. We believe Humanoid Everyday will serve as a valuable resource for advancing versatile and intelligent humanoid robotic agents.

While Humanoid Everyday provides a comprehensive dataset for humanoid manipulation, we only evaluate existing imitation learning policy architectures. Although these baselines provide useful insights, their performance degrades on more challenging tasks due to the high dimensionality of humanoid action spaces, indicating the need for more specialized model designs. In addition, our cloud-based evaluation system does not yet support automatic scene resetting, as current imitation learning policies are not sufficiently robust for humanoids to recover the environments without human assistance. In future work, we plan to develop more robust humanoid policies and extend our evaluation system to support autonomous scene recovery.
\section{Acknowledgements}
The USC Physical Super Intelligence Lab acknowledges generous supports from Toyota Research Institute, Dolby, Google DeepMind, Capital One, Nvidia, and Qualcomm. This work was partially supported by the National Science Foundation through NSF CPS \#2434460. Yue Wang is also supported by a Powell Research Award.
% \jiageng{We may need a user study here showing that our proposed evaluation platform is better than traditional methods, or we refer to other cloud evaluation papers to see how they evaluate their manipulation policies. We need some experimental evaluation here.}

\bibliographystyle{IEEEtran}
\bibliography{custom}

\end{document}